\documentclass[]{article}

\usepackage{graphicx}        
\usepackage{multicol}        
\usepackage[framemethod=TikZ]{mdframed}  
\usepackage{amsmath}
\usepackage{amssymb}
\usepackage{pifont}
\usepackage{rotating}
\usepackage{multirow}
\usepackage{subfigure}
\usepackage{xcolor}
\usepackage{colortbl}
\usepackage{fancyvrb}
\usepackage{mathptmx}
\usepackage{helvet}
\usepackage{courier}
\usepackage{varwidth}
\usepackage[noend]{algpseudocode}
\usepackage[affil-it]{authblk}

\usepackage{geometry}
\newtheorem{definition}{Definition}

\newcommand{\HTN}{HTN}
\newcommand{\tHTN}{state-based HTN }

\newcommand{\pHTN}{plan-based HTN }

\newcommand{\noah}{NOAH}

\newcommand{\ie}{i.e.,}

\newcolumntype{x}[1]{>{\centering\let\newline\\\arraybackslash\hspace{0pt}}p{#1}}
\newcolumntype{C}[1]{>{\centering\let\newline\\\arraybackslash\hspace{0pt}}m{#1}}
\newcolumntype{L}[1]{>{\raggedright\let\newline\\\arraybackslash\hspace{0pt}}m{#1}}
\newcolumntype{Y}{>{\centering \arraybackslash} X }

\def\qed{\ifmmode{\scriptsize
    QED}\else{\unskip\nobreak\hfil\penalty50\hskip1em\null\nobreak\hfil{\scriptsize
      QED}\parfillskip=0pt\finalhyphendemerits=0\endgraf}\fi\medskip}

\mdfdefinestyle{MyFrame}{%
    linecolor=gray!100!white,
    outerlinewidth=1pt,
    roundcorner=8pt,
    innertopmargin=8pt,
    innerbottommargin=8pt,
    innerrightmargin=8pt,
    innerleftmargin=8pt,
    backgroundcolor=gray!10!white}
\newcounter{examplecounter}
\newenvironment{myExample}{\vspace*{4mm}\begin{mdframed}[style=MyFrame]%
  \refstepcounter{examplecounter}%
  \quad
  }{%
  \end{mdframed}%
\vspace*{4mm}
}

\begin{document}

\title{Introduction to AI Planning}

\author{Marco Aiello and Ilche Georgievski}
\affil{Service Computing Department, University of Stuttgart \\ firstname.lastname@iaas.uni-stuttgart.de}

\maketitle

\begin{abstract}
    These are notes for lectures presented at the University of Stuttgart that provide an introduction to key concepts and techniques in AI Planning. \textit{Artificial Intelligence Planning}, also known as \textit{Automated Planning}, emerged somewhere in 1966 from the need to give autonomy to a wheeled robot. Since then, it has evolved into a flourishing research and development discipline, often associated with scheduling. Over the decades, various approaches to planning have been developed with characteristics that make them appropriate for specific tasks and applications. Most approaches represent the world as a state within a state transition system; then the planning problem becomes that of searching a path in the state space from the current state to one which satisfies the goals of the user. 
    
    The notes begin by introducing the state model and move on to exploring classical planning, the foundational form of planning, and present fundamental algorithms for solving such problems. Subsequently, we examine planning as a constraint satisfaction problem, outlining the mapping process and describing an approach to solve such problems. The most extensive section is dedicated to Hierarchical Task Network (HTN) planning, one of the most widely used and powerful planning techniques in the field. The lecture notes end with a bonus chapter on the Planning Domain Definition (PDDL) Language, the de facto standard syntax for representing non-hierarchical planning problems.
\end{abstract}

\vspace{8cm}

\noindent\copyright 2024 Marco Aiello \& Ilche Georgievski. All rights reserved. This work is posted on arXiv under a non-exclusive license to distribute. Permission for reuse beyond personal purposes must be obtained from the authors.

\newpage

\tableofcontents

\newpage

\section{State Model}
\label{sec1}

To achieve goal-oriented behaviour, AI Planning systems choose available actions to change the state of the environment in order to satisfy the goal of their user. In this sense, the { state model} is a fitting representation. It is a standard model in AI defined over a state space, that is, a set of states and a set of actions that deterministically map each state to another one, extended with a single initial state and a non-empty set of goal states~\cite{nilsson1980:aiprinciples}. If systems employing the state model have complete knowledge about the states, then the state model is {\em fully observable}. The concept of a state model is also the one that most planning approaches rely on.

A {\em state model} $\mathcal{M}$ consists of five components:
\begin{itemize}
\item $S$ is a finite set of states,
\item $s_{0}\in S$ is the initial state,	
\item $S_{G}\subseteq S$ is the set of goal states,
\item $A$ is a finite set of actions,
\item $\delta:S\times A\rightarrow S$ is a deterministic transition function.
\end{itemize}

$\mathcal{M}$ may be represented by a labelled directed graph whose vertices are the states in $S$. The graph contains an edge from $s$ to $s'$ if and only if  $s'\in\delta(s,a)$, with $a\in A$. Each arc is called a {\em state transition.}

\begin{figure}[htb]
\begin{center}
\includegraphics[width=69mm]{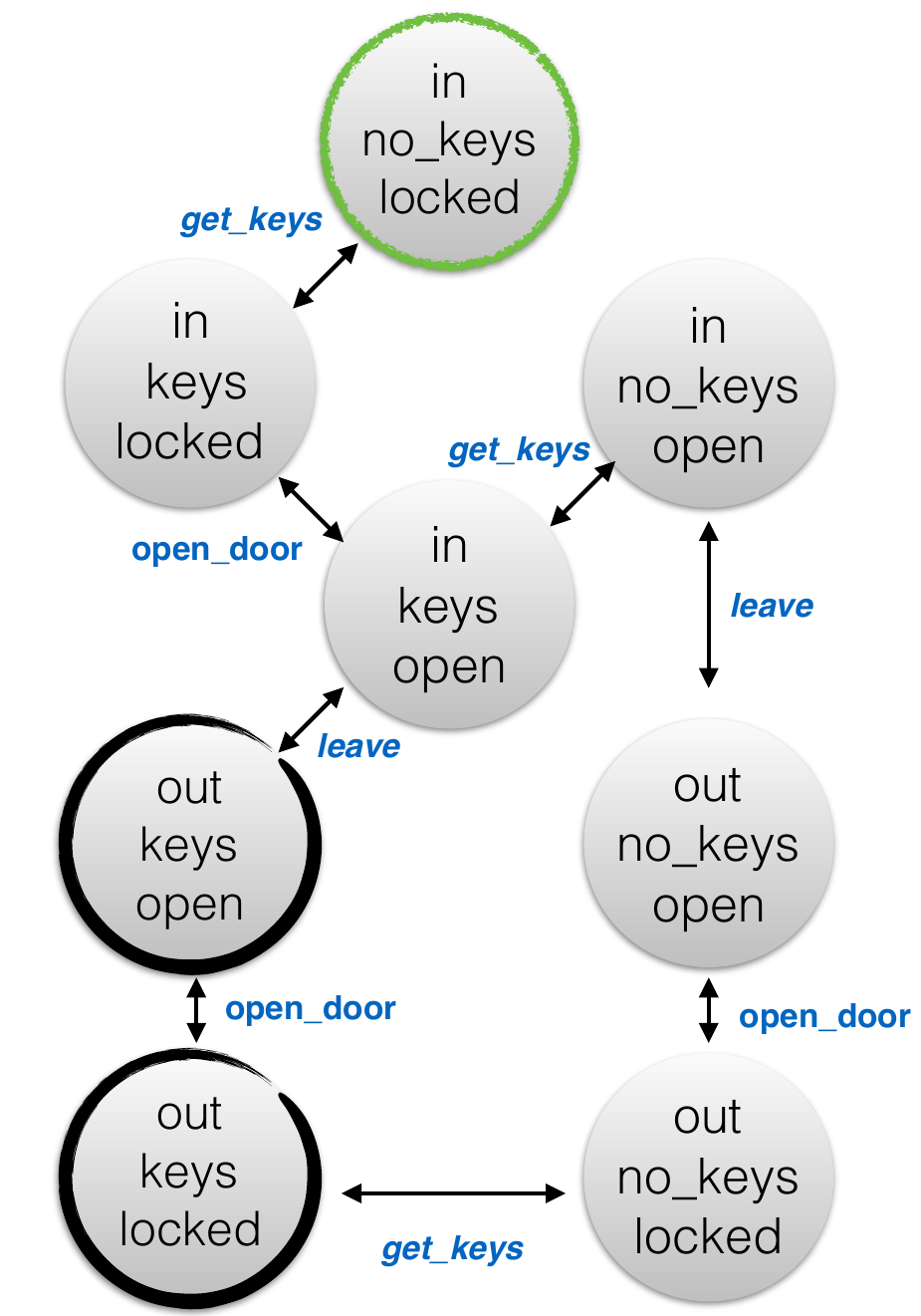}
\end{center}
\caption{An example of a state model.\label{ch2aip:fig:ex1}}
\end{figure}

\newpage

\begin{myExample}
{\bf Example.} Consider a simple pervasive computing scenario where there is a set of keys with an active tag, a person and a home with a door. The person can be inside or outside the home. The door can be closed and locked or open, and the keys can be in the possession of the person or not. A state model for this example is the following one.
\begin{itemize}
\item $S=\{(in, no\_keys, locked), (in, keys, locked),\ldots, (out, no\_keys, open)\}$ in other terms, the states are all possible combinations of three boolean variables: $|S|=2^3$
\item $s_{0}=(in, no\_keys, locked)$ is the initial state,	
\item $S_{G}=\{(out,keys,locked),(out,keys,open)\}$ is the set of two goal states,
\item $A=\{get\_keys, open\_door, leave, drop\_keys, close\_door, enter, lock\_door, \linebreak unlock\_door\}$ is the set of actions,
\item $\delta:\{(in, keys, locked)\cdot open\_door\to(in, keys, open),\ldots\}$ is the transition function. The full set of functions and states is shown in a graphical form in Figure~\ref{ch2aip:fig:ex1}. The starting state is circled in green, while the two goal states are circled in thick black. Furthermore the transtions are indicated by a double-headed arrow representing both the action and its inverse, e.g., $get\_keys$ and $drop\_keys$.\\[1.2ex]
The figure shows only arrows for the actions in $\delta$. The actions shown in italic are human actions, i.e., enter/leave, get/drop\_keys while all others are device operations. For each action, its inverse is also in $\delta$, e.g., $get\_keys^{-1}=drop\_keys$, therefore the double-headed arrows. Parallel actions are not possible, that means that one cannot go from $(in, keys, open)\to(out, no\_keys, open)$ as this would entail both dropping the keys and leaving the home in one state transition.
\end{itemize}
\end{myExample}

The state model $\mathcal{M}$ is {\em deterministic}, meaning that for every pair of a state $s$ and an action $a$, $\delta(s,a)$ is either empty or returns a single state. An action $a$ is {\em applicable} to state $s$ if $\delta(s,a)$ returns a state. Thus, the application of $a$ to a state $s$ results in a deterministic system being in the state $s'=\delta(s,a)$, or equivalently $s'=s[a]$. The system stays in the same state until another action is applied. This implies that no events are allowed to change the state of the system outside of the set of actions defined in $\mathcal{M}$. We then say that the state model is {\em static}.

Once a state model is defined, it is often the practice to formulate a `solution' or `plan' for such a state model. While the state model may be intrinsically linked to some problem, it does not require a solution of any kind. What can be actually described is a trajectory of states that, through actions, leads from an initial state to a final state that is also a goal state. Such a trajectory may then correspond to a solution to any problem defined over the state model.

Given an initial state $s_0$, a sequence of states $s_{0},\dots,s_{n+1}$ is called a {\em solving trajectory} if and only if there exists an action $a_{i}$ such that $s_{i+1}=s_{i}[a_{i}]$, and $s_{n+1}\in S_{G}$.

\begin{myExample}
{\bf Example.} The action $open$ is applicable to state $(in, keys, locked)$, as the user has the keys and the door is locked. Other actions are not applicable, e.g., it is not possible to open the door without the keys, transition $(in, no\_keys, locked)\to (in, no\_keys, open)$. A path of applicable actions from the initial state to a final state, that is, a solution is, for example, $s_0[get\_keys,\linebreak open\_door,leave]$.
\end{myExample}

The simplest problem of planning for a state model $\mathcal{M}$ is to choose which actions to apply to which states in order to transition from the given initial state to one of the goal states. A {\em solution} to such a problem is a sequence of actions $a_{1},\dots,a_{n}$ corresponding to a solving trajectory in $\mathcal{M}$. The structure holding the solution to the problem is called a {\em plan}, often denoted by $\pi$.

State models for a particular domain are usually built by explicitly defining the entire state space, that is, enumerating all possible states and state transitions. This becomes infeasible in practice for most real applications as the size and complexity of state models increase. A better approach is to define the state models implicitly by using a representation in which a state consists of a collection of variables. The actions, their applicability, and the transition function are then defined in terms of these variables.

\section{Classical Planning}
\label{sec3}

The state model $\mathcal{M}$ introduced in Section~\ref{sec1} implies complete knowledge, that is, the model is {\em fully observable}. The application of an action in a given state always leads to one state only -- the model is {\em deterministic}. Finally, a state cannot be changed by any dynamics other than the actions given in the state model -- the model is {\em static}. Planning for a state model that is fully observable, deterministic, and static is called {\em classical planning}. The term {\em classical planning} refers to planning approaches that restrict the view of the world over which planning is performed. As a result, classical planning relies on several assumptions that simplify the state model~\cite{ghallab2004:automated}. These are: 
\begin{itemize}
\item environments have a finite set of states, 
\item the initial state is complete and fully observable, 
\item actions are deterministic, 
\item environment states change only by executing actions, 
\item goals are either satisfied or not by plans, 
\item plans are ordered sequences of actions, and 
\item there is no explicit representation of time.
\end{itemize}

In contrast, non-classic planning approaches relax these assumptions and incorporate a wider range of properties describing actual environments. For example, modern planning approaches are able to deal with states that are not fully observable or with actions whose application can non-deterministically bring a system to several possible states. Therefore, much work has focused on developing planning approaches that relax one or more of these assumptions by allowing incomplete knowledge about the initial state, partially observable states, non-deterministic actions, actions with conditional effects, preferences, and extended goals, partially ordered plans, durative actions, and so on.

\begin{myExample}
{\bf Historical Note.} Shakey, the name of the first known developed robot, used various programs for perception, modelling of the environment, and problem-solving. The main component used for problem-solving is the STRIPS planner, which is an automated planner developed in 1971 at the Stanford Research Institute. The planner is considered one of the first planning algorithms and representations~\cite{fikes1971:strips}. The name of the planner is derived from the STanford Research Institute Problem Solver. STRIPS actions for Shakey involved moving from one location to another, changing the robot's direction, blocking and unblocking doors with objects, and pushing movable objects to particular locations.

The STRIPS representation, or STRIPS language, is used to refer to the formal language describing the input to the planner. The original version of STRIPS language was based on first-order logic, which provided for serious challenges in relation to the interpretation of the STRIPS operator descriptions~\cite{lifshitz1987:strips-semantics}. The language was later simplified and reduced to propositional logic~\cite{nilsson1980:aiprinciples}.  \end{myExample}

A common way of defining state models underlying classical planning problems is by using the STRIPS language. We use the version of STRIPS representation as assumed in the definition of the STRIPS subset of PDDL rather than the original version to define a classical planning problem.

\begin{definition}[Classical planning problem]\label{def:planning-problem}
A classical planning problem $\mathcal{P}$ is a tuple $\langle F,O,I,G \rangle$, where
\begin{itemize}
\item $F$ is a set of atoms,
\item $O$ is a set of operators each of which is of the form $\langle pre(o),add(o),del(o)\rangle$, where $pre(o)$ are preconditions, $add(o)$ and $del(o)$ are add and delete lists, respectively, and $pre(o),add(o),del(o)\subseteq F$,
\item $I\subseteq F$ is the initial state, 
\item $G\subseteq F$ is the goal state. 
\end{itemize}
\end{definition} 

Notice that now the initial state is a set of atoms defining a unique state in the domain, while the goal state is the one for which the set of atoms in $G$ are true. 

The state model underlying the classical planning problem is implicitly described by the STRIPS representation. More formally, a classical planning problem $\mathcal{P}=\langle F,O,I,G \rangle$ implicitly describes a state model $\mathcal{M}_\mathcal{P}=\langle S,s_{0},S_{G},A,\delta \rangle$, where
\begin{itemize}
\item the states $s\in S$ are collections of atoms from $F$,
\item the initial state $s_{0}$ is $I$,
\item the goal states $s\in S_{G}$ are such that $G\subseteq s$,
\item the actions $a\in A$ are the operators $o\in O$ such that $pre(o)\subseteq s$ for $\delta(s,a)$,
\item the transition function $\delta$ maps states $s$ into states $s'=(s\cup add(a))\setminus del(a)$ for $a\in A$.
\end{itemize}

Before we provide a definition of the solution to a classical planning problem, we define the application of a sequence of operators $o_{1},\dots,o_{n}$ to a state $s$ as
\begin{align*}
s[] &= s \\
s[o_{1},\dots,o_{n}] &= (s[o_{1},\dots,o_{n-1}])[o_{n}]
\end{align*}

\begin{myExample}\label{pp:classicalplanning}
{\bf Example.} Take the example presented in Figure~\ref{ch2aip:fig:ex1} and let us now model it as a classical planning problem. We have that:
\begin{itemize}
\item $F=\{Inside(agent), Locked(door), Owns(key)\}$;
\item $O=\{getKeys =\langle \lnot Owns(key), Owns(key), \lnot Owns(key)\rangle,\\
 \mbox{}\quad unlock=\langle \{Locked(door),Owns(key)\},\lnot  Locked(door) ,  Locked(door)\rangle,\\
  \mbox{}\quad exit=\langle  Inside(agent), \lnot Inside(agent), Inside(agent)\rangle, \ldots\}$,
\item $I=\{Inside(agent),\lnot Owns(key),Locked(door)\}$, and
\item $G=\{\lnot Inside(agent),Owns(key)\}$.
\end{itemize}
While a plan now becomes $\pi=getKeys, unlock, exit$. We notice that the preconditions to the $getKeys$ actions are met by the initial state of the model, while $unlock$ cannot act on the initial state, but it can on the outcome of the successful application of $getKeys$. More precisely, the application of $getKeys$ to the initial state $I$, means the addition of owning the keys, making the state $\{Inside(agent),Owns(key),Locked(door)\}$, now the preconditions of $unlock$ are met and it can be applied, bringing the state to $\{Inside(agent),Owns(key), \linebreak \lnot Locked(door)\}$. Finally, the preconditions of $exit$ are met, bringing the agent to a subset of the goal state $\{\lnot Inside(agent),Owns(key),\lnot Locked(door)\}$. 
\end{myExample}

\newpage

Given a classical planning problem, we are now in the position to define a solution to it.
\begin{definition}[Plan]\label{def:solution} 
Let $\mathcal{P}=\langle F,O,I,G \rangle$ be a classical planning problem. The sequence of operators $\pi=a_{1},\dots,a_{n}$ is a plan for $\mathcal{P}$ if each operator $o_{i}$ is applicable in $o_{i}$, that is, $pre(o_{i})\subseteq s_{i}$, and the state resulting from the application of $\pi$ from the initial state $I$ contains the goal state $G$, that is, $G\subseteq s[\pi]$.
\end{definition}

The plans for the problem $\mathcal{P}$ correspond to the solving trajectories of the state model $\mathcal{M}_{\mathcal{P}}$.

In practical applications, the domains can be quite large and complex. An example of application is from the logistics domain, particularly multi-modal transportation, where at least two means for transport are combined in order to move goods~\cite{florez2011:multi-modal-transportation}. Consider that there is a truck used to move a container from one location to another. Before the move happens, the truck should be loaded, for which a load action is defined that specifies the truck to be used, the container to be loaded on that truck, and the location where the loading happens. If we assume that there are 250 trucks, 250 containers, and 720 locations, then there are $250\times 250\times 720=45$ million possible action instances only for the load action. Similar situations can be encountered in other applications, such as the Deep Space Mission of NASA, the game of bridge, video games, and so on. Searching for solutions to planning problems in such domains can be very expensive, if not unfeasible. It is important to know, at least from a theoretical point of view, the complexity of tasks related to planning in their various forms.

\subsection{Complexity of Classical Planning}
\label{ch2:sec3:sub1}

The worst-case complexity is a common way to get a general understanding of the feasibility and difficulty of solving a computational problem~\cite{papadimitriou2003computational}. In our case, a representative problem to consider is that of finding a plan, if it exists. More formally, given a planning problem $\mathcal{P}$, the {\em plan existence problem} or \texttt{PlanEx($\mathcal{P}$)} consists in finding whether there exists a plan that solves $\mathcal{P}$. Given a planning problem $\mathcal{P}$ and an integer $k$, the {\em bounded plan existence problem} or \texttt{PlanLen($\mathcal{P}$)} consists of finding whether there is a plan of length at most $k$. \texttt{PlanEx($\mathcal{P}$)} and \texttt{PlanLen($\mathcal{P}$)} help in understanding the plan generation problem for $\mathcal{P}$ and its optimality in terms of number of actions in the plan, respectively.

For classical planning, both plan existence problems are decidable. This means that we can construct an algorithm that checks whether a solution exists for any possible planning problem in the category of classical planning problems. This is guaranteed by the fact that the number of states is finite. However, if we add function symbols to the representation of planning problems, then the number of states can be infinite, and \texttt{PlanEx($\mathcal{P}$)} may become semidecidable. This means that we can construct an algorithm that terminates when a solution exists, but it may not terminate on planning problems that do not have a plan.

In addition to knowing whether and in which cases the plan existence problems are solvable, one typically wants insight into how difficult it is to solve them or what resources are needed. In computational complexity, $\mathbf{PSPACE}$ defines a class of complex problems that can be solved by a deterministic Turing machine using a polynomial amount of space. The space here refers to the computer memory used by the computation. Problems with both plans' existence fall into this category. Even more strongly, they are, in fact, $\mathbf{PSPACE}$-complete~\cite{bylander1994:stripscomplexity}. A decision problem is $\mathbf{PSPACE}$-complete if (i) it is using a polynomial amount of space (that is, it is in $\mathbf{PSPACE}$) and (ii) every other decision problem in $\mathbf{PSPACE}$ can be transformed to it in polynomial time. In other words, $\mathbf{PSPACE}$-complete problems can be solved in constant space and time. Note that $\mathbf{PSPACE}$-complete problems are the most difficult problems in $\mathbf{PSPACE}$. Even if we reduce the complexity of planning problems by making restrictions on the operators, the decision problems remain difficult to solve in the worst case. For example, if we restrict operators to be without delete lists in effects, both problems become $\mathbf{NP}$-hard. If we though remove negative preconditions, then \texttt{PlanEx($\mathcal{P}$)} can be solved in polynomial time.

Such strong complexity results should, however, not discourage us. These are worst-case complexity and might say very little of the average case or the one most likely to occur for problems in a specific domain. The planning community typically also evaluates approaches in terms of their practical performance, often on a set of benchmark domains, regardless of the theoretical worst-case results.

\subsection{Planning Algorithms}

Planning is the general process of going from the planning problem to its solution. The planner operates on a search space, with a search strategy, going through a potentially very large space or even the whole space in the worst case looking for solutions. The search space can be of several forms and structures, where the difficulty of the search increases with the intricacy and size of the space.

There are many ways to find solutions to planning problems. The simplest way of solving planning problems is by using search algorithms whose search space is a subset of the state space. These are called {\em state-based search algorithms.} We briefly look at two state-based search algorithms, namely forward search and backward search. 

\subsubsection*{Forward search}
Forward search starts from the initial state and searches forward through the space of states in an attempt to find the goal state. Fig.~\ref{ch2:alg:forward-search} shows a forward search algorithm that can be used to solve a planning problem $\mathcal{P}$. The algorithm is both sound, which means a plan returned by it is a solution to $\mathcal{P}$, and complete, which means that if there is a solution to $\mathcal{P}$, the algorithm will find it.

\begin{figure}
\noindent\fbox{%
\begin{varwidth}{\dimexpr\textwidth-2\fboxsep-2\fboxrule\relax}
\begin{algorithmic} 
\Function{Forward-search}{$\mathcal{P}=\langle O$, $s_{0}$, $g\rangle$}
\State $s\leftarrow s_{0}$
\State $\pi\leftarrow$ an empty plan
\Loop
\If{$s$ satisfies $g$} {\Return $\pi$}
\EndIf
\State $\mathit{applicableActions}\leftarrow$ \{$a\:|\:a$ is a ground instance of an operation in $O$, and $pre(a)$ is satisfied in $s$\}
\If{$\mathit{applicableActions}=\emptyset$} {\Return failure}
\EndIf
\State non-deterministically choose an action $a\in\mathit{applicableActions}$
\State $s\leftarrow\delta(s,a)$
\State append $a$ to $\pi$
\EndLoop
\EndFunction
\end{algorithmic}
\end{varwidth}%
}
\caption{A forward-search planning algorithm. It takes a planning problem $\mathcal{P}$ and looks for a sequence of actions that would solve the problem. It terminates when a solution is found or when no solutions exist. The algorithm is identical to the one presented in ~\cite{ghallab2004:automated}.}\label{ch2:alg:forward-search}
\end{figure}

Forward search is regarded as too inefficient in practice. There are two reasons for this, the first of which is that forward search also explores irrelevant actions. Consider the \verb|get-keys| action and the goal \verb|(holds k1)|. Now suppose that there are 10 keys in total the user owns, all given in the initial state. So, we can bind 10 different values for the parameter \verb|?k| in \verb|get-keys|, leading to 10 possible instances of actions. The forward-search algorithm would need to try all these actions to find the one that leads to the goal.

As second, planning problems often have large search spaces. Even for small planning problems, the state space may consist of an exponential number of states (see~\cite{russell2003:modern} for an example).

Forward search becomes feasible in practice by employing heuristics that can lead the search faster to the goal state. Today, heuristics are derived automatically from the representation of planning problems. 

\newpage

\begin{myExample}
{\bf Example.} We apply the algorithm in Figure~\ref{ch2:alg:forward-search} to the example in Figure~\ref{ch2aip:fig:ex1}. A possible trace of the execution where the first applicable action is always chosen first
\begin{enumerate}
\item $s=s_0=\{in,no\_keys,locked\}$
\item $\pi=\emptyset$
\item $applicableActions = \{get\_keys\}$
\item $s = \{in,keys,locked\}$
\item $\pi = get\_keys$
\item $applicableActions = \{open\_door,drop\_keys\}$ 
\item $s = \{in,keys,open\}$
\item $\pi = get\_keys,open\_door$
\item $applicableActions = \{leave,drop\_keys,lock\_door\}$
\item $s = \{out,keys,open\}$
\item $\pi = get\_keys,open\_door,leave$
\item terminate as the condition of $s$ being a goal state is met and return $\pi$
\end{enumerate}

\medskip
\noindent
In this example, we have assumed that the first non-deterministically chosen action was always contributing to brining us to the solution. If that were not the case, one would get a failure, entailing that the non-deterministic choice would fall on the second applicable action, and so on, until all applicable actions were explored. In the end, a solution would be found if it existed due to the finite state assumption.
\end{myExample}

\subsubsection*{Backward search}

Backward search starts from a goal state and applies actions backward until it finds the sequence of actions leading up to the initial state. To apply an action backward means to remove the action's added effects from the current goal state and add the action's preconditions to the current goal state. Assuming that the action was executed, its add effects need not be true before that state, while preconditions must have held for the action to have been applicable. We do not use delete effects because it is impossible to know whether they are true or not before an action's application; it is only known that they are no longer true.

Backward search chooses actions that are relevant to the goal state (or current goal state). An action is relevant to a goal if some of its effects unify with some part of the goal. The effects of a relevant action must not negate any part of the goal. Fig.~\ref{ch2:alg:backward-search} shows the backward search algorithm, which is also sound and complete.

\begin{figure}
\noindent\fbox{%
\begin{varwidth}{\dimexpr\textwidth-2\fboxsep-2\fboxrule\relax}
\begin{algorithmic} 
\Function{Backward-search}{$\mathcal{P}=\langle O$, $s_{0}$, $g\rangle$}
\State $\pi\leftarrow$ an empty plan
\Loop
\If{$s_{0}$ satisfies $g$} {\Return $\pi$}
\EndIf
\State $\mathit{relevantActions}\leftarrow$ \{$a\:|\:a$ is a ground instance of an operator in $O$ that is relevant for $g$\}
\If{$\mathit{relevantActions}=\emptyset$} {\Return failure}
\EndIf
\State non-deterministically choose an action $a\in\mathit{relevantActions}$
\State prepend $a$ to $\pi$
\State $g\leftarrow\delta^{-1}(g,a)$
\EndLoop
\EndFunction
\end{algorithmic}
\end{varwidth}%
}
\caption{The non-deterministic algorithm for backward-search planning, from~\cite{ghallab2004:automated}.}\label{ch2:alg:backward-search}
\end{figure}

In backward search, the goal \verb|(holds k1)| is immediately unified with the effect \verb|(holds ?k)| of the action \verb|get-keys|, meaning that the parameter \verb|?k| is bound to the object \verb|k1|. We would then go backward using the \verb|get-keys| action to create the predecessor state \verb|(owns a1 k1)|, which is part of the initial state, so the search is finished.

In summary and in contrast to forward search, which also goes through irrelevant actions, backward search considers only relevant actions and thus usually has a smaller search space. It does, however, use sets of states instead of individual states, which are used by forward search, for which case it is difficult to design good heuristics. This is the main reason why forward search is chosen more often than backward search.

\section{Planning as CSP}
\label{sec4}

The original planning state model $\mathcal{M}$ consists of states that are units with no particular internal structure, see Section~\ref{sec1}. In the following section, we explored how to use variables as building blocks of predicates which are in turn used to represent states and classical planning problems. It is then natural to turn to the case in which variables are direct constituents of states. A state is represented by a set of variables, each of which takes some value. The problem then becomes the identification of appropriate values to assign to variables given some constraints over allowable values. This constitutes the {\em constraint satisfaction problem}, or CSP. A solution to a CSP is a set of assignments to variables such that all constraints are satisfied.

\begin{myExample}
{\bf Constraint satisfaction.} In constraint satisfaction, one can naturally express a wide variety of computational problems that deal with mappings and assignments. Constraint satisfaction originated in the 1970s independently in Artificial Intelligence, Database Theory, and Graph Theory. It was later realized, in the 1990s, that these streams of research were, in fact, different perspectives of the same fundamental problem. Today, constraint satisfaction concepts are used for defining algorithms in theoretical computer science, and for modelling and solving real-world problems in AI, such as planning and scheduling, and in applied computer science, such as natural language comprehension.

The concept of a constraint, which was probably borrowed from other research fields (see~\cite{vanemden2006:CSPreview}), was a central notion used to describe relationships or dependencies between variables in terms of their possible values. Many techniques designed and introduced earlier, such as backtracking and a form of constraint propagation, seemed to be relevant to dealing with variables and the dependencies between them. As a then new discipline, constraint satisfaction developed these ideas further and introduced new topics, such as local consistency search. The main focus of constraint satisfaction is on developing new forms of search intertwined with consistency techniques and introducing new types of constraints.

A popular approach of the 1980s was the implementation of constraint satisfaction by embedding constraints directly into a host programming language. This led to the development of a new discipline called {\em constraint programming}. Constraint programming languages have libraries that provide constraint propagation and various forms of search. For the principles of constraint programming, see~\cite{apt2003:cp}.
\end{myExample}

The idea is to solve planning problems by constraint satisfaction. To that end, a planning problem needs to be mapped to a CSP in such a way that there is a guarantee that the solution to the CSP is also a solution to the original planning problem. %

\subsection{CSP}

A constraint satisfaction problem $\mathcal{P_{CSP}}$ is made of three components:

\begin{itemize}
\item $X=\{x_{1},\dots,x_{n}\}$ is a finite set of variables,
\item $D=\{D_{1}\dots,D_{n}\}$ is a set of finite domains such that there is one domain for each variable,
\item $C$ is a set of constraints that restrict the values the variables can take.
\end{itemize}

A {\em domain} $D_{i}$ of variable $x_{i}$ is a set of possible values $\{v_{1},\dots,v_{k}\}$ that can be assigned to $x_{i}$. A {\em constraint} $c_{i}$ is a pair $\langle t,R\rangle$, where $t$ is a tuple of variables (called {\em scope}), and $R$ is a relation that defines the values the variables can take. A constraint can be specified explicitly by listing all tuples of allowed values or implicitly by using a relation symbol. Consider the two variables $x_{1}$ and $x_{2}$ having the same domain $\{\alpha,\beta\}$; then to define that, for example, these variables must have different values, a constraint can be specified as $\langle(x_{1},x_{2}),[(\alpha,\beta),(\beta,\alpha)]\rangle$ or $\langle(x_{1},x_{2}),x_{1}\ne x_{2}\rangle$.

In CSP, a state consists of a set of pairs $(x_{i},v_{i})$ representing an assignment of values to some or all of the variables in $X$. If the assignment has values for all variables, it is called a {\em complete} assignment. An assignment that satisfies all constraints is called a {\em consistent} assignment.  A {\em solution} to a CSP is a complete and consistent assignment. A CSP is {\em consistent} if such a solution exists.

\subsection{From Planning Problems to CSPs}

We are now ready to encode a classical planning problem into a CSP. A common practice is to translate a planning problem whose plans have a length of at most $k$, a number given a priori. Such a problem is called a {\em bounded} planning problem. For the sake of clarity, we assume that classical planning problems are represented by variables with binary values instead of predicates. We refer to such variables as {\em state variables}. This means that for every ground atom, there is a state variable with a domain $D$ of two values: true or false. Similarly, a state variable is ground if there is a value assigned to it. The representation based on state variables enables us to explain the encoding into a CSP compactly.

Our final aim is to characterise sequences of states $\{s_{0},\dots,s_{k}\}$ that correspond to plans of a length at most $k$ steps. In the following, we refer to a state $s_{i}$ by its index $i$, for $0\leq i\leq k$. We begin the encoding of a classical bounded planning problem $P$ based on state-variable representation into a constraint satisfaction problem $P_{CSP}$ by defining the variables of $P_{CSP}$. The CSP variables are all ground state variables of $P$ augmented with one variable whose value corresponds to the action applied in state $i$.
\begin{itemize}\label{ch2:alg:cspEncoding}
\item For each ground state variable $f$ with a domain $D$ and for each $0\leq i\leq k$, there is a CSP variable $x[i]$, whose domain is $D$. 
\item For each $0\leq i\leq k-1$, there is a single CSP variable $a[i]$, whose domain is the set of possible actions.
\end{itemize}

\medskip\noindent
Once the CSP variables are derived, the next step is to encode the initial state, the goal state and action into constraints.
\begin{itemize}
\item Constraints encoding $s_{0}$: Every ground state variable $f$ whose value is $true$ in $s_{0}$ is encoded into a constraint of the corresponding CSP variable for $i=0$ of the form $(x[0]= true)$. Every ground state variable $f$ not mentioned in $s_{0}$ is encoded into the constraint $(x[0]=false)$.
\item Constraints encoding $g$: Every ground state variable $f$ whose value is $v$ in $g$ is encoded into a constraint of the corresponding CSP variable for $i=k$: $(x[k]=v)$.
\item Constraints encoding actions: For each ground action (an action instance of some operator $o\in O$) $a$ and for each $0\leq i\leq k-1$:
	\begin{itemize}
	\item Constraints encoding preconditions: Every atom in $pre(a)$ can be represented as a condition of the form $(x=v)$, which is encoded into a constraint $(a[i]=a,x[i]=v)$.
	\item Constraints encoding effects: Every atom in the add list can be represented as an assignment of a state variable $f\leftarrow true$, which is encoded into a constraint: $(a[i]=a, x[i+1]=true)$. For every atom in the delete list, there is an assignment of a state variable $f\leftarrow false$, which is encoded into a constraint: $(a[i]=a, x[i+1]=false)$.
	\end{itemize}
\end{itemize}

\begin{myExample}
{\bf Example.} 
To encode the exiting the room example as a CSP, we first need to set a bound on the length of sought plans, say four steps, $k=4$. This means that we have three vectors of four elements for the three ground variables $in[i],$ $keys[i]$, $open[i]$ all ranging over boolean values $\{T,F\}$. Given the bound to four actions per plan, we have four variables $a[i]$ ranging over the domain of actions $D_A=\{get\_keys,$ $open\_door,$ $leave,$ $drop\_keys,$ $close\_door,$ $enter,$ $lock\_door,$ $unlock\_door\}$. 

In the initial state, the agent has no keys, and the door is closed, resulting in the following constraints $in[0]=T$ and $keys[1]=open[2]=F$.

The goal is to be out with the keys, this means $in[4]=F$ and $keys[4]=T$.

Next consider the $get\_keys$ action with precondition $\lnot Owns(key)$, effect $Owns(key)$, and deletion of $\lnot Owns(key)$. This becomes encoded in the following constraint for the precondition ($a[i]=get\_keys$, $keys[i]=F$), while the effect becomes ($a[i]=get\_keys$, $keys[i+1]=T$).
Two frame axioms for this case become ($a[i]=get\_keys$, $in[i]=in[i+1]$, $open[i]=open[i+1]$). The encoding of the other actions follows a similar procedure.

\end{myExample}

Since not all state variables are affected by the effects of selected actions, there are variables that remain unchanged from one step to the next one and which need to be explicitly asserted. This is accomplished by frame axioms. For every action $a$ and state variable $f$ with value $v$ ($true$ if the corresponding ground atom is in the state or $false$ otherwise) that is not affected by $a$, a frame axiom is encoded as: $\{(a[i]=a,x[i]=v,x[i+1]=v)\}$.

Now that we have a bounded classical planning problem encoded into a CSP, $P_{CSP}$ can be solved. If a solution exists, the outcome will be an assignment of values to all CSP variables. If variables $a[i]$ get the assignment $a[i]=a_{i+1}$ for $0\leq i\leq k-1$, then each $a_{i}$ is an action of $P$, and the sequence $a_{1},\dots,a_{k}$ is a plan of the bounded planning problem.

Note that the state-variable representation of the bounded classical planning problem can be easily generalised to multi-valued variables, which are more common in real-world applications. In this case, the state variables have finite domains of values that can be encoded into CSP variables with the corresponding multi-valued domains. The approach of encoding the initial state, the goal state, and actions follows the steps used for encoding the binary-valued representation.

\subsection{Complexity}

Solving constraint satisfaction problems is in general $\mathbf{NP}$-complete, as the size of $P_{CSP}$ can be exponential, that is, $\Pi^{i=m}_{i=1}\:|D_{i}\:|$. Several restricted forms of CSP have been investigated to find tractable classes of problems. There are two main and well-studied types of restrictions, namely the constraint language restriction and the structural restriction. The former restricts the constraint language in terms of the available types of constraints: the set of relations on a domain in a given $P_{CSP}$ must be fixed and finite. Thus, the constraint language restricts the domain and the set of relations of each constraint. If, for each fixed constraint language, there exists a polynomial algorithm that solves all inconsistencies, then the corresponding of $P_{CSP}$ is tractable (in $\mathbf{P}$). Otherwise, $P_{CSP}$ is $\mathbf{NP}$-complete.

The structural restriction limits the way constraints are placed over variables: the immediate interaction between variables in a $P_{CSP}$ is bounded. Thus, the structural restriction focuses on the scopes of the constraints rather than their relations. The problems based on structural restrictions are tractable.

Encoding a bounded classical planning problem into a constraint satisfaction one, the number of CSP variables $m$ is, in fact, linear in the size of the planning problem, $k(n+1)$, where $k$ is the bound of the plan length and $n$ is the number of state variables. From Section~\ref{ch2:sec3:sub1}, we know that the bounded decision problem \textsc{PlanLength} for a classical planning problem is $\mathbf{PSPACE}$-complete.

\subsection{Algorithms}

A CSP is usually solved by searching a variable assignment from several possibilities. The search is performed systematically by a backtracking algorithm that checks all possible assignments of values to variables. A basic backtracking algorithm works as follows. It incrementally expands a partial consistent solution to a complete one by repeatedly choosing an unassigned variable and trying all values in its domain in such a way the chosen value is consistent with the current partial solution. Given that the variables are assigned sequentially, the validity of a constraint is checked as soon as the scope of the constraint is assigned. If a constraint is violated in a partial solution, the search backtracks by reconsidering the most recently assigned variables for which there are still unexplored values. In doing so, backtracking guarantees to find a solution if it exists. Otherwise, the problem has no solution. The drawback of backtracking is efficiency, having to check all possibilities, in turn, leading to an unpractical runtime for large search spaces, given the exponential complexity. In our specific planning case, after encoding the planning problem as a CSP, the algorithm starts with the initial state $s_0$ rather than the empty solution and proceeds forward action by action till it reaches the bound. This is illustrated in Figure~\ref{ch2:alg:csp}. One could add a condition in the loop that if a consistent 

\begin{figure}
\noindent\fbox{%
\begin{varwidth}{\dimexpr\textwidth-2\fboxsep-2\fboxrule\relax}
\begin{algorithmic} 
\Function{Constraint satisfaction}{$\mathcal{P}=\langle g,S,s_0,O\rangle$,$k$}
\State Encode $\mathcal{P}$ into $\mathcal{P_{CSP}}$ of length $k$ (see page~\pageref{ch2:alg:cspEncoding})
\State $i=0$
\State $Solution\leftarrow s_0$
\For{$i<k$ and $Solution$ is consistent}
\State Non-deterministically select $a[i]$ and assign value $action$
\State $Solution$ add $(a[i]=action)\bigcup\{$all of $action$'s pre and post conditions$\}$
\If{$g$ is satisfied}
\State Return $a[0],a[1],\ldots a[i]$
\EndIf
\EndFor
\State Return Failure (no plan found)
\EndFunction
\end{algorithmic}
\end{varwidth}%
}
\caption{The algorithm for solving CSP encoded planning problems.}\label{ch2:alg:csp}
\end{figure}

The attentive reader may notice that this basic backtracking strategy identifies assignment conflicts rather late. Since it assigns values sequentially, it might detect a conflict only after it assigns values to all variables of a conflicting constraint. To achieve earlier detection, techniques for consistency checks are commonly embedded within backtracking algorithms.

Constraint propagation is the process of checking the consistency of values of variables, which in turn allows the reduction of the number of possible values for a variable. This reduction can influence the set of possible values of other variables, thus reduction can be propagated and applied to the whole sets of variables.

When embedded within backtracking algorithms, constraint propagation can be applied after the backtracking algorithm assigns a value. In this case, a consistency check is performed on the unassigned ``future'' variables. For each unassigned variable related to the one just assigned, one removes any value in the domain of the unassigned variable that is inconsistent with the value of the current variable. This technique of consistency checking is called {\em forward checking}. Forward checking enables the backtracking algorithms to prune early the parts of the search space that do not lead toward a solution.

Forward checking detects inconsistencies between the currently assigned variable and the future variables, thus it makes only the current variable consistent. A more general approach is to look ahead more broadly to all variables by checking the constraints between all future variables. Look ahead enables the backtracking algorithm to prune the search space earlier than forward checking.

Whenever a new variable is assigned, all its current values are guaranteed to be consistent with past variables by both techniques. On the other hand, both techniques are computationally expensive in terms of time. 

Other classes of search algorithms can also be considered to solve CSPs. Rather than systematically searching through all possible alternatives (therefore, paths from an initial state), local search algorithms explore a single current state and move only to its successor states without retaining information on the past path. In the context of CSP, local search algorithms use an initial state in which a value is assigned to every variable while the search progresses by choosing a new value of one variable at a time. When a new value is chosen, that value should produce a minimum number of conflicts with other variables. This is called a min-conflict heuristic. The two main benefits of local search algorithms are the limited use of memory space and the finding of reasonable solutions in large state spaces.

\section{HTN Planning}
\label{sec5}

We saw that the classical form of planning requires an initial state, a goal state, and some actions to realise a sequence of actions that, when executed in the initial state, lead to the goal state. While actions represent simple transitions from one world state to another one, a very common structure we use to understand the world better is of a hierarchical nature. The ability of planning to represent and deal with hierarchies is at the heart of {\em hierarchical planning}, or more specifically {\em Hierarchical Task Network (HTN) planning}~\cite{georgievski2015:htn}. Hierarchies encompass rich domain knowledge characterising the world, which makes \HTN~planning very useful, and also to perform well in real-world domains.

\HTN~planning breaks with the tradition of classical planning due to the type of domain knowledge and the type of goal it requires. The basic idea includes an initial state, an initial task network as an objective to be accomplished, and domain knowledge consisting of networks of primitive and compound tasks. A task network represents a hierarchy of tasks each of which can be executed, if the task is primitive, or decomposed into refined subtasks. The planning process starts by decomposing the initial task network and continues until all compound tasks are decomposed, that is, a solution is found. The solution is a plan which equates to a sequence of primitive tasks applicable to the initial state. 

\begin{myExample}
{\bf Historical Note.}
Beside being a tradition breaker, \HTN~planning appears to be controversial as well. The controversy lies in its requirement for well-conceived and well-structured domain knowledge. Such knowledge is likely to contain rich information and guidance on how to solve a planning problem, thus encoding more of the solution than was envisioned for classical planning techniques. This structured and rich knowledge gives a primary advantage to \HTN~planners in terms of speed and scalability when applied to real-world problems and compared to their counterparts in classical world. 

The biggest contribution towards this kind of ``popular'' image of \HTN~planning has emerged after the proposal of the Simple Hierarchical Ordered Planner (SHOP)~\cite{nau1999:shop} and its successors. SHOP is an \HTN-based planner that shows efficient performance even on complex problems, but at the expense of providing well-written and possibly algorithmic-like domain knowledge. Several situations may confirm our observation, but the most well known is the disqualification of SHOP from the International Planning Competition (IPC) in 2000~\cite{bacchus2001:aips} with the reason that the domain knowledge was not well written so that the planner produced plans that were not solutions to the competition problems~\cite{nau1999:shop}. Furthermore, the disqualification was followed by a dispute on whether providing such knowledge to a planner should be considered as ``cheating'' in the world of AI planning~\cite{nau2007:trends}.

SHOP's style of \HTN~planning was introduced by the end of 1990s, but \HTN~planning existed long before that. The initial idea of hierarchical planning was presented  in 1975 as the Nets of Action Hierarchies (\noah) planner~\cite{sacerdoti1975:structure}. It was followed by a series of studies on practical implementations and theoretical contributions on \HTN~planning up until today. We believe that the fruitful ideas and scientific contribution of nearly 40 years must not be easily reduced to controversy and antagonism towards \HTN~planning.
\end{myExample}

There are several proposals to formalise a model for HTN planning . Each proposal defines hierarchical concepts appropriate to its underlying theory. In order 

To set the grounds for basic understanding of HTN planning, we provide a hierarchical planning model in which we keep definitions of constructs generic. We complement the generic definitions by specifics as needed throughout the book. For other more specific or slightly different in style models of hierarchical planning, we refer the reader to, for example,~\cite{erol1994:umcp,nau1999:shop,ghallab2004:automated,geier2011:decidability}.

As the original STRIPS planning language, the \HTN~planning language is a first-order language, with well-defined semantics. Its basic construct is a predicate which the reader should be already familiar with (see Section~\ref{sec2}). Here too, the state $s$ is a set of ground predicates in which the closed-world assumption is adopted.

Characteristic for HTN planning are the notions of primitive and compound tasks. A {\em primitive task} is an expression $t_{p}(\tau)$, where $t_{p}$ is some primitive symbol and $\tau=\tau_{1},\dots,\tau_{k}$ are constant symbols and/or variables. The set of all primitive tasks is a finite one.

Each primitive task is represented by a single operator defined similarly as for the classical planning problem. An {\em operator} $o$ is a triple $(p(o),\mathit{pre(o)},\mathit{eff(o)})$, where $p(o)$ is a primitive task, $\mathit{pre(o)}$ are preconditions, $\mathit{eff(o)}$ are effects. The subsets $\mathit{pre}^{+}(o)$ and $\mathit{pre}^{-}(o)$ denote positive and negative preconditions of $o$, respectively, and $\mathit{eff}^{-}(o)$ and $\mathit{eff}^{+}(o)$ are negative and positive effects of $o$, respectively.

Similarly to the classical planning problem, a transition from a state to another one is accomplished by an instance of an operator whose preconditions are a logical consequence of the current state. That is, an operator $o$ is {\em applicable} in state $s$, if $\mathit{pre}^{+}(o)\subseteq s$ and $\mathit{pre}^{-}(o)\cap s=\emptyset$. The application of $o$ to $s$ results in state $s[o]=(s\cup \mathit{eff}^{+}(o))\setminus \mathit{eff}^{-}(o)=s'$.

What makes HTN planning different from classical planning and a unique planning technique is the domain knowledge expressed through compound tasks. In a similar way to primitive tasks, a {\em compound task} is an expression $t_{c}(\tau)$, where $t_{c}$ is a compound symbol and $\tau = \tau_{1},\dots,\tau_{k}$ are constant symbols and/or variables. The set of compound tasks is also a finite state. We refer to the union of the sets of primitive and compound tasks as a set of tasks $T$. 

Each compound task is associated with one or more methods holding the domain knowledge. A {\em method} $m$ is a pair $\langle c(m),tn(m)\rangle$, where $c(m)$ is a compound task, and $tn(m)$ is a task network. A {\em task network} $tn$ is a pair $\langle T',\psi\rangle$, where $T'$ is a finite set of tasks, and $\psi$ is a set of constraints. Constraints specify restrictions over $T'$ that must be satisfied during the planning process and by the solution. We refer to a task network over the set of primitive tasks as a {\em primitive task network}. 

We can now define the notion of planning problem for HTN planning.

\begin{definition}[\HTN~planning problem]\ignorespaces\label{def:pp}
An {\em \HTN~planning problem} $P_{HTN}$ is a tuple $\langle Q,T,O,M,tn_{0},s_{0}\rangle$, where
\begin{itemize}
\item $Q$ is the finite set of predicates,
\item $T$ is the finite set of tasks,
\item $O$ is a finite set of operators,
\item $M$ is a finite set of methods,
\item $tn_{0}$ is an initial task network,
\item $s_{0}$ is the initial state.
\end{itemize}
\end{definition}

From this definition, we should understand that planning is no longer searching for a sequence of actions that maps an initial state into some goal state, but instead searching for a sequence of actions that accomplishes the initial task network when applied to the initial state. 

Recall that an operator sequence $o_{1},\dots,o_{n}$ is applicable in state $s$ if there is a sequence of states $s_{0},\dots,s_{n}$ (that is, a solving trajectory) such that $s_{0}=s$ and $o_{i}$ is applicable in $s_{i-1}$ and $s_{i-1}[o_{i}]=s_{i}$ for all $0 \leq i \leq n$. Then, given an HTN planning problem $P_{HTN}$, a plan is a solution to $P_{HTN}$ if there exists an operator sequence applicable in $s_{0}$ by decomposing $tn_{0}$. What exactly means to decompose a task network depends on various factors described in the following.

\subsection{Concepts}

\begin{figure} 
	\centering
	\includegraphics[width=\textwidth]{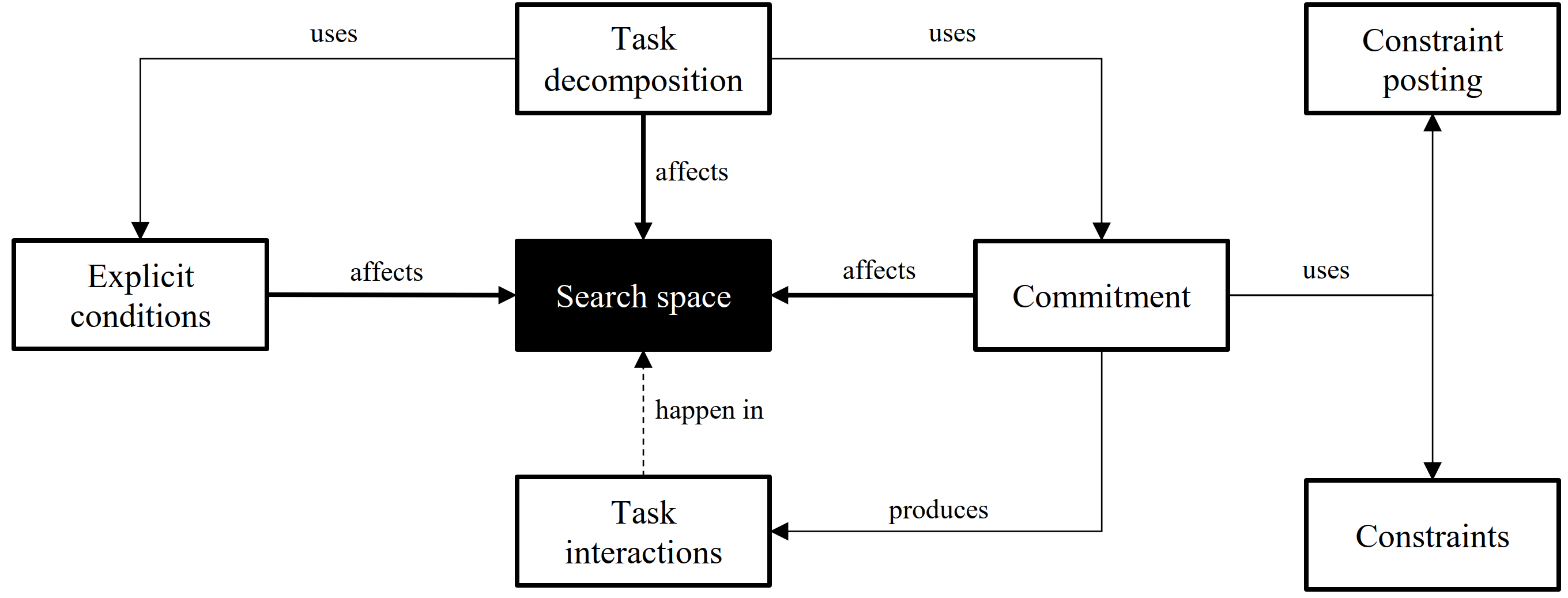}
	\caption{Conceptual model of \HTN~planning.}\label{fig:concepts}
\end{figure}

To go beyond the basic hierarchical model and understand the workings of \HTN~planning, one has to look into the concepts characterising and influencing the planning process and decisions made along. Figure~\ref{fig:concepts} shows a diagram of the conceptual model of \HTN~planning. A key concept of the model is the search space to which other concepts are related and interconnected in various ways. Let us shed some light into each of them.

\subsubsection{Task decomposition}\label{sec:td}
Simply put, a task decomposition is a strategy that transforms task networks. To be more specific, given a task network $tn$, a {\em task decomposition} chooses a task $t$ from $tn$ and, if $t$ is primitive and applicable to the current state $s$, the task decomposition applies $t$ to $s$. Otherwise, the decomposition strategy analyses all the methods that contain $t$ as a part of their definition. Assuming that a set of methods is found, the task decomposition makes a non-deterministic choice of a method $m$, and replaces $t$ with the task network associated with $m$. Finally, the task decomposition checks the newly composed task network against any constraint-related violation and modifies it, if necessary.

Task decomposition can be characterised by the representation of task networks in terms of task ordering, and the way of forming new task networks during decomposition. As a consequence, there are three styles of task decomposition: {\em totally ordered task decomposition} (TOTD), {\em unordered task decomposition} (UTD), and {\em partially ordered task decomposition} (POTD). The first one follows the assumption of total order on task networks so as when a task is decomposed, the new task network is created in such a way that newly added tasks are totally ordered among each other and with respect to the tasks of the existing task network. Sometimes we refer to \HTN~planning that uses TOTD as totally ordered \HTN~planning. UTD relaxes the requirement of totally ordered task networks in such a way that tasks can be totally ordered or unordered with respect to each other, however no tasks in parallel are allowed. When a task is decomposed, new task networks are created in a manner that newly added tasks are interleaved with the tasks of the existing task network until all permissible permutations are exhausted. In this case, we refer to \HTN~planning as unordered \HTN~planning. The last style allows the existence of a partial order on tasks. When a task is decomposed, the tasks in the newly created network can be ordered in parallel whenever possible, certainly with respect to the constraints. We refer to \HTN~planning that uses POTD as partially ordered \HTN~planning.

\subsubsection{Search space}\label{sec:ss}
There are two structures of the search space in hierarchical planing. The first one consists of task networks and task decompositions as evolutions from one task network to another. At the beginning of the search, a task decomposition is imposed on the initial task network of a given \HTN~planning problem. The planning process continues by repeatedly decomposing tasks from the updated task network until a primitive task network is produced. In other words, the initial task network is reduced to a primitive task network. Interestingly, the task network can be seen as a partially specified plan up until the point in search where the task network is primitive and fully specified. A linearisation of such a primitive task network, which is also applicable in the initial state, represents a solution to the given planning problem.

\begin{figure} 
	\centering
	\includegraphics[width=.5\textwidth]{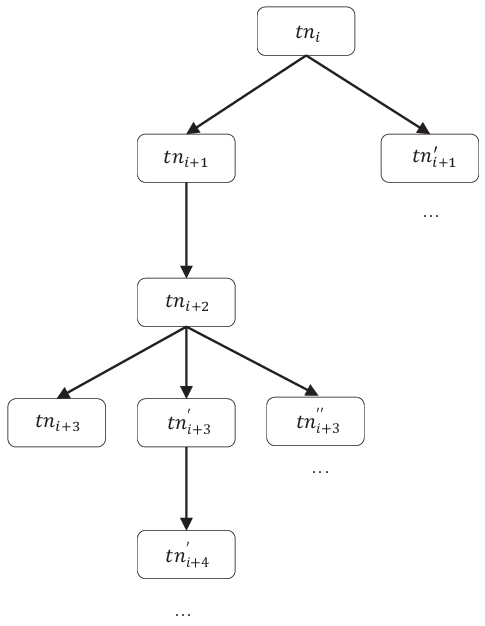}
	\caption{Plan space represented as a directed graph with task networks as vertices and task decompositions as edges.}\label{fig:pss}
\end{figure}

It should be evident that the search space is a directed graph in which task networks (or partial plans) are vertices, and a decomposition of one task network into another task network by some method is an outgoing edge, under the condition that the initial task network belongs to the graph. An excerpt of such a graph is shown in Figure~\ref{fig:pss}. Due to vertices being partial plans, we use the term {\em plan space} to refer to this structure of search space, and to the model of \HTN~planning that employs a plan space as \pHTN planning. Formal definition of plan space follows.

\begin{definition}[Plan space]
	Given an \HTN~planning problem $P_{HTN}$, a {\em plan space} $PG$ is a directed graph $(V,E)$ such that $tn_{0} \in V$, and for each $tn\rightarrow_{D}tn'$: $tn,tn' \in V$ and $(tn,tn') \in E$.
\end{definition}
 
The second space structure is in essence a subset of state space. It consists of explicitly described states restricted by task decompositions. As in classical planning, the search begins in the initial state with an empty plan, but instead of searching for a state that will satisfy the goal state, the search is for a state that will accomplish the initial task network. In particular, if a task from the task network is compound, the task decomposition continues on the next decomposition level, but in the same state. Otherwise, the task is executed and the search continues into a successor state. The task in the latter case is then added to the plan. When there are no more tasks in the task network to be decomposed, the search is finished. The solution to the given \HTN~planning problem is the plan containing a sequence of (totally ordered) primitive tasks.

\begin{figure} 
	\centering
	\includegraphics[width=\textwidth]{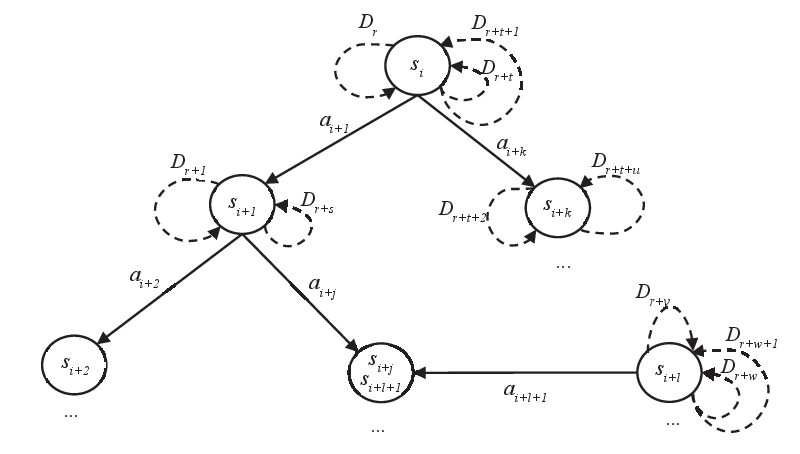}
	\caption{State space as a directed graph with task networks as vertices and task decompositions as edges.}\label{fig:sss}
\end{figure}

We consider a state space as a directed graph in which a state is a vertex, and a task decomposition maps to the same state where the corresponding method is applicable, and an operator application leads to a successor state. An excerpt of such a graph is shown in Figure~\ref{fig:sss}. For the obvious reasons, we use the term {\em state space} to refer to this structure of search space, and to the model of \HTN~planning employing a state space as \tHTN planning. Formal definition of state space follows.

\begin{definition}[State space]
	Given an \HTN~planning problem $P_{HTN}$, a {\em state space} $SG$ is a directed graph $(V,E)$ such that $s_{0} \in V$, and there is a state $s_{i}$ and $t_{k} \in tn$ such that
	\begin{itemize}
		\item if $t_{k}$ is primitive, then $s_{i}[t_{k}]=s_{i+1}$ such that $k=i+1$, $s_{i},s_{i+1} \in V$ and $(s_{i},s_{i+1}) \in E$; or
		\item if $t_{k}$ is compound, then $tn\rightarrow_{D}tn'$ is a self-transition such that $s_{i} \in V$ and $(s_{i},s_{i}) \in E$.
	\end{itemize}
\end{definition}

\subsubsection{Constraints}\label{sec:constraints}
As we have seen in Section~\ref{sec4}, constraints express relationships between two or more variables. In the context of \HTN~planning, task networks rely upon such constraints as per definition, and also upon constraints that can be added dynamically during planning to resolve inconsistencies, for example. 

For the purpose of hierarchical planning, there are three types of constraints\footnote{These are in fact similar to the three interpretations of constraints in~\cite{stefik1981:constraints}.}. The first type of constraints implies commitments about partial descriptions of state objects. It answers the question of whether to make some object more specific and how specific by binding values to one or more variables describing that object. This type of constraints is directly related to the commitment strategy underlying the planning process, as described in the following section. Another type of constraints refines variable bindings in the sense that it eliminates a certain variable binding if that binding does not satisfy some given condition. The last type of constraints expresses relations between variables in different parts of a task network. What this means is that variables can be shared between different part of the task network, which is the same as saying variables can be shared among various steps in the (partial) plan, in the case of \pHTN~planning.

\subsubsection{Commitment}
Hierarchical planning is similar to other planning techniques in the sense that it needs to make two decisions on constraints. The first one is the common decision on constraints for binding variables, while the second one is on ordering constraints, in this case, ordering tasks in task networks. 

There are two main approaches for when and how to make decisions on constraints. The first approach manages constraints in compliance with the {\em least-commitment strategy} by which task ordering and variable bindings are deferred until a decision is forced~\cite{weld1994:least_commitment}. The least-commitment strategy allows for partially described state objects as some variables of a given object may be bound, while other may not.

The second approach handles constraints according to the {\em early-commitment strategy} by which all object variables are bound and operators in the plan are totally ordered at each step of the planning process. Planners employing this strategy greatly benefit from the possibility of adopting forward chaining in which chaining of operators is achieved by imposing a total order over the plan. The total ordering ensures that neither the current operator to be added to the plan can interfere with some earlier operator's preconditions or effects, nor a later operator can interfere with current task's preconditions or effects. 

\subsubsection{Constraint posting}\label{sec:constraint-posting}
The concept of manipulating constraints in task networks is known as {\em constraint posting}. It is based on well-known operations on constraints, being the main ones: constraint satisfaction, constraint propagation, and constraint formulation. Constraint satisfaction ensures that a variable binding satisfies some given constraints, as in CSP described in Section~\ref{sec4}, or guarantees the consistency of a set of ordering constraints over tasks in a given task network.

Constraint propagation enables adding or retracting constraints to and from task networks. Variable constraints in one part of a task network can be propagated based on variable constraints in another part of that task network. Ordering constraints are propagated during the so-called linking process. When some task interferes with another task, the {\em linking process} records a causal link -- a 3-tuple of two pointers to tasks $t_{e}$ and $t_{p}$, and a predicate $q$ which is both an effect of $t_{e}$ and a precondition of $t_{p}$ -- and adds it to the task network.

The last operation is different from the other two in that it does not happen during planning. Constraint formulation is the process of modelling constraints within the domain knowledge, especially when a domain author is aware of some possible impasse situations in advance. By posting constraints as control information into the domain knowledge, planners can gain on efficiency by refining the search space~\cite{erol1994:semanticshtn, nareyek2005:constraints}.

Constraint posting is in fact the underlying process of addressing task interactions, elsewhere known as conflict resolution~\cite{yang1992:conflict} or critics~\cite{tate1976:project,wilkins1988:extending}.

\subsubsection{Task interactions}
A {\em task interaction} is a connection between two tasks (or parts) of a task network in which these tasks (or parts) have a certain effect on each other. The character of this effect yields two categories, namely harmful interactions and helpful interactions. {\em Harmful interactions}, also knowns as threats or flaws, are defined as relationships that introduce conflicts among different parts of a task network that threaten network's correctness. The most common harmful interactions are the following:
\begin{itemize}
	\item {\em Deleted-condition interaction} happens when a primitive task in one part of a task network deletes an expression that is a precondition of a primitive task in another part of that task network.
	\item {\em Double-cross interaction} appears when an effect of each of two conjunctive primitive tasks deletes a precondition for the other. That is, an effect of the first task deletes a precondition of the second primitive task, and an effect of the second task deletes a precondition of the first task.
	\item {\em Resource interaction} occurs in two situations, and it is subdivided accordingly. A \textit{resource-resource interaction} is similar to the deleted-condition interaction, while a \textit{resource-argument interaction} occurs when a resource in one part of a task network is used as an argument in another part of that task network.
\end{itemize}

{\em Helpful interactions} are relationships that can be found in situations when one part of a task network makes use of information associated with another part in the same task network. The detection of these interactions implies a possibility for planners to generate task networks and solutions of better quality. That is, some tasks can be merged together, eliminating task redundancy and potentially optimising the cost of the solution~\cite{foulser1992:merging}. The most common helpful interactions are the following: 

\begin{itemize}
	\item {\em Placeholder replacement} appears when an actual value already exists for a particular formal object. We already know that \HTN~planning allows tasks with variables to be inserted into a task network. If there is no specific value to be chosen for a particular variable, a so-called formal object is created and bound to that variable~\cite{sacerdoti1975:nonlinear}. The formal object is simply a placeholder for something unspecified at that point.
	\item {\em Phantomisation} emerges when some fact, which is supposed to be achieved, is already true at the point in the task network where it occurs. In other words, a phantomisation of a task $t$ with an effect $q$ is considered accomplished by treating $q$ as achieved and finding an existing task $t'$ in the task network that already achieved the same effect $q$.
	\item {\em Disjunct optimisation} happens in disjunctive goals when one disjunctive goal is ``superior to the others by the nature of its interaction'' with the other tasks in a task network~\cite{sacerdoti1975:nonlinear}.
\end{itemize}

The reader might have recognised that the task interactions are an inevitable consequence of the commitment strategy chosen.

\subsubsection{Explicit conditions}\label{sec:ec}
Hierarchical planners essentially depend on the quality of domain knowledge so as to restrict and guide the search for a solution. Domain authors are undoubtedly the ones who have the responsibility of giving guidance information. Domain authors explicitly encode such information as conditions within tasks. For example, explicit conditions may restrict the task decomposition, filter applicable methods, expect phantomisation, or require information from an external resource. The following is a list of types of explicit conditions that can be encountered in the existing \HTN~planners.

\begin{itemize}
	\item {\em Supervised condition} is accomplished within a compound task. The condition may be satisfied either by an intentional insertion of a relevant effect earlier in the processing of a task network, or by an explicit introduction of a primitive task that will achieve the required effect. Only this condition should allow further decompositions to be made.
	\item {\em External condition} must be accomplished at the required task, but under the assumption that it is satisfied by some other task from the same task network.
	\item {\em Filter condition} decides on task relevance to a particular situation. In the case of method relevance to a certain task decomposition, this condition reduces the branching factor by eliminating inappropriate methods.
	\item {\em Query condition} accomplishes queries about variable bindings or restrictions at some required point in a task network.
	\item {\em Compute condition} requires satisfaction by information coming only from external systems, such as a database.
	\item {\em Achieve condition} allows expressing goals that can be achieved by any means available to a planner.
\end{itemize}

\subsection{Complexity}
Here too, we are interested in deciding whether a plan exists. For the purpose of analysing the complexity of solving an \HTN~planning problem, we provide a few more clarifications and assumptions. The words and phrases in italics are used as categories in the analysis.

\begin{itemize}
\item The sets $O$ and $M$ can be provided in two ways. The sets can be a part of the {\em input}, or they can be {\em fixed} in advance, meaning the tasks are allowed to contain methods corresponding only to predicates in the initial state.
\item A compound task can be defined in several ways: (1) a compound task is without any restriction ({\em yes}); (2) a {\em regular} task in task networks -- at most one compound task followed by all primitive tasks; (3) an {\em acyclic} task -- a task can be decomposed to only a finite depth; and (4) compound tasks are not allowed at all ({\em no}).
\item A task network containing primitive and compound tasks as defined in the previous point can be {\em totally ordered} or {\em partially ordered}.
\item Variables can be allowed or not in $P_{HTN}$.
\end{itemize}

The complexity results of HTN planning are summarised in Table~\ref{tab:complx}. When no restrictions on compound tasks are imposed and task networks are partially ordered, then giving $O$ and $M$ in the input or fixing them in advance, or allowing variables or not, does not affect the outcome and \textsc{PlanEx} is undecidable. However, given $O$ and $M$ in the input, and being every task acyclic and every task network partially ordered, \textsc{PlanEx} becomes decidable. \textsc{PlanEx} is decidable when task networks are totally ordered. In particular, when unrestricted compound tasks and variables are allowed, \textsc{PlanEx} is $\mathbf{EXPSPACE}$-hard in double exponential time (2-EXPTIME), or, if no variable is allowed, \textsc{PlanEx} is $\mathbf{PSPACE}$-hard in exponential time. When only primitive tasks and variables are allowed, \textsc{PlanEx} is $\mathbf{NP}$-complete, irrespective of the ordering of task networks. Furthermore, forbidding the use of variables in the same case makes \textsc{PlanEx} to be in $\mathbf{P}$. However, disallowing variables when task networks are partially ordered tasks does not change the outcome and \textsc{PlanEx} remains $\mathbf{NP}$-complete. Regardless of the ordering of task networks, when compound tasks are regular, there are two outcomes. When $O$ and $M$ are given in the input, and if variables are allowed, then \textsc{PlanEx} is $\mathbf{EXPSPACE}$-complete, otherwise \textsc{PlanEx} is $\mathbf{PSPACE}$-complete. When $O$ and $M$ are fixed in advance, and variables are allowed, \textsc{PlanEx} is $\mathbf{PSPACE}$-complete.

\begin{table*} \footnotesize
\caption{Computational complexity of \HTN~planning (adapted from~\cite{erol1995:complexity})}\label{tab:complx}
\centering
\begin{tabular}{llllr}
\hline
{\bf Operators and methods} & {\bf Compound task} & {\bf Task network} & {\bf Variables} & {\bf Plan existence}\\ \hline
fixed & \multirow{7}{*}{yes} & \multirow{3}{*}{partially ordered} & \multirow{2}{*}{yes} & undecidable\\ \cline{1-1} \cline{5-5}
\multirow{14}{*}{input} & & & & undecidable\\ \cline{4-5}
& & & no & undecidable \\ \cline{3-5}
& & \multirow{6}{*}{totally ordered} & \multirow{2}{*}{yes} & 2-EXPTIME \\
& & & & EXPSPACE-hard \\ \cline{4-5}
& & & \multirow{2}{*}{no} & EXPTIME \\ 
& & & & PSPACE-hard \\ \cline{2-2} \cline{4-5}
& \multirow{5}{*}{no} & & yes & NP-complete \\ \cline{4-5}
& & & no & P \\ \cline{3-5}
& & \multirow{4}{*}{partially ordered} & yes & NP-complete\\ \cline{4-5}
& & & no & NP-complete \\ \cline{2-2} \cline{4-5}
& \multirow{2}{*}{acyclic} & & yes & decidable \\ \cline{4-5}
& & & no & decidable \\ \cline{2-5}
& \multirow{4}{*}{regular} & \multirow{4}{*}{unimportant} & yes & EXPSPACE-complete\\ \cline{4-5}
& & & no & PSPACE-complete \\ \cline{1-1} \cline{4-5}
\multirow{2}{*}{fixed} & & & \multirow{2}{*}{yes} & PSPACE\\
& & & & PSPACE-complete\\ 
\end{tabular}
\end{table*}

\section{Bonus: PDDL}
\label{sec2}

A standard representation for describing problems in planning is the Planning Domain Definition Language (PDDL). PDDL, as a first-order representation, is built around objects and relations. Looking at the example in Figure~\ref{ch2aip:fig:ex1}, `agent', `door', and `key' are objects, and the verb phrases `is inside', 'is locked' and `owns' refer to relations. The PDDL representation consists of predicates which are statements describing objects with certain relations among them and can be either {\em true} or {\em false}. Predicates can be also referred to as {\em atoms} or {\em facts}. An atom is composed of a predicate symbol, which refers to a relation, followed by a list of arguments, which are used to refer to objects. For part of the aforementioned example, we get the atom $Inside(agent)$. Arguments can be constant symbols or variables. In order to determine whether a predicate is true or false, an interpretation is needed of exactly which objects and relations are referred to by arguments and predicate symbols. In our example, $Inside(agent)$ states that the agent is inside the smart home. Note that there may be many possible interpretations for a single predicate. We say that a predicate is true when the relation denoted by the predicate symbol holds among the objects denoted by the arguments. A predicate that cannot be shown to be true is considered to be false. Such conclusion is known as the {\em closed-world assumption}, that is, everything that is not explicitly stated to have a certain value, remains the same. A predicate whose arguments contain no variables, that is, all variables are bound to objects, is called a {\em  atom}. We say that a parametrised predicate {\em unifies} with a ground atom when the parameters of the predicate can be substituted with the argument values of the atom. For example, $Inside(?a)$ unifies with $Inside(agent)$ with $?a$ bound to $agent$.

As in other first-order languages, complex statements can be formulated from predicates by using logical connectives, including negation (denoted by \verb|not|), conjunction (denoted by \verb|and|), disjunction (denoted by \verb|or|), implication (denoted by \verb|imply|), universal quantification (denoted by \verb|forall|), and existential quantification (denoted by \verb|exists|). For the semantics of the logical connectives, see for instance~\cite{russell2003:modern}.

By using predicates, PDDL enables describing actions, initial states, and goals. PDDL separates the descriptions of predicates and actions from the description of an initial state and a goal. The former is referred to as a domain description (or domain definition), while the latter as problem description (or problem definition). A planning problem is then created by combining a domain description with a problem description. A consequence of this separation is the possibility to create different planning problems in one domain by using the same domain description combined with many different problem descriptions.

The syntax of PDDL is based on Lisp\footnote{Lisp is a functional programming language with parenthesized prefix notation, introduced in the late 50s by John McCarthy.}, making a description of a planning problem to be structured as a list of parenthesized expressions.

In the following, we explain in more details the syntax and structure of a domain description and a problem description, we introduce a set of additional features that may increase the expressiveness of the language or that may give a more compact syntax. Finally, we introduce the STRIPS subset of the PDDL language.

\begin{myExample}
{\bf Historical Note.} The Planning Domain Definition Language (PPDL) was initially created to provide a common syntax for specifying planning problems for the (first) International Planning competition in 1998. The original PDDL was based on function-free first-order logic, meaning that functions and arithmetics were not included in the logic, thus terms could only be constants or bound variables (and analogously for preconditions). The language has been further extended to support better expression of aspects of real-world planning problems, including numbers, numeric expressions and action durations, predicates derived from other predicates, and plan constraints and preferences. For details on the different versions of PDDL, see~\cite{mcdermott1998:pddl,fox2003:pddl21,edelkamp2004:pddl22,gerevini2006:pddl3}. 
\end{myExample}

\subsection{Domain Description}
A domain description is expressed by keywords and special fields that start with a colon. Each domain description begins with the declaration

\medskip
\verb|(define (domain <name>))|,

\medskip\noindent 
where \verb|define| and \verb|domain| are keywords, and \verb|<name>| defines the name of the application domain. For example, a domain description for a smart home can begin with the declaration 

\medskip
\verb|(define (domain smart-home))|.

\medskip
\noindent In the simplest domain description, the declaration of the domain name is followed by the definitions of predicates and actions. The \verb|:predicates| contains a list of declarations of the predicates. A predicate is of the form 

\medskip
\verb|(NAME ?A1 ... ?An)|, 

\medskip
\noindent where the arguments beginning with a question mark are parameters that represent objects in the domain. Consider the predicates in Fig.~\ref{fig:pddl-simple-predicates}. The predicate \verb|(agent ?a)| can be used to describe that \verb|?a| is an agent, while \verb|(in ?a ?r)| to determine whether \verb|?a|, that is, an agent is in \verb|?r|, which is a room. The interpretation of the latter predicate depends on the combinations of values for the parameters for which the predicate can be true and the relationship to other predicates, such as \verb|(agent ?a)| and \verb|(room ?r)|. Similar reasoning can be applied for the other predicates in Fig.~\ref{fig:pddl-simple-predicates}.

\begin{figure}[!ht]
\begin{tabular}{l}
\verb|(:predicates|\\
\hspace{0.5cm}\verb|(agent ?a)|\\
\hspace{0.5cm}\verb|(room ?r)|\\
\hspace{0.5cm}\verb|(door ?d)|\\
\hspace{0.5cm}\verb|(key ?k)|\\
\hspace{0.5cm}\verb|(in ?a ?r)|\\
\hspace{0.5cm}\verb|(owns ?a ?k)|\\
\hspace{0.5cm}\verb|(adjacent ?r1 ?r2)|\\
\hspace{0.5cm}\verb|(installed-in ?d ?r)|\\
\hspace{0.5cm}\verb|(in ?a ?r)|\\
\verb|)|
\end{tabular}
\caption{PDDL predicates representing some of the properties for the key example in Figure~\ref{ch2aip:fig:ex1}.}\label{fig:pddl-simple-predicates}
\end{figure}

The Boolean values of predicates determine whether the predicates are facts holding in a given state of an environment. The truth values of predicates are affected by the actions, whereby the predicates are usually not influenced by a single action only. In PDDL, actions are schemas consisting of three elements, namely parameters, preconditions and effects. An action description begins with the declaration \verb|:action <name>|, where \verb|<name>| is the name of the action schema. Each of the elements of an action schema has its own special field. Fig.~\ref{fig:pddl-simple-action} shows an action schema for opening a door in a smart home as described in our key example. 

\begin{figure}[!ht]
\begin{tabular}{l}
\verb|(:action open-door|\\
\hspace{0.5cm} \verb|:parameters (?d)|\\
\hspace{0.5cm} \verb|:precondition (and (door ?d) (not (opened ?d)))|\\
\hspace{0.5cm} \verb|:effect (opened ?d)|\\
\verb|)|
\end{tabular}
\caption{PDDL action schema for a door operation that opens a door.}\label{fig:pddl-simple-action}
\end{figure}

The action name and parameters define what the action can accomplish. Each action schema characterises a set of possible action instances by using the parameters. A specific action or action instance is derived by assigning each of the parameters a value (or object) that comes from a specific problem instance. An action instance is also called ground action. The preconditions define the conditions that need to be satisfied before the action can be applied. Preconditions are usually a complex statement, meaning they are expressed as a logical formula over predicates using logical connectives. In Fig.~\ref{fig:pddl-simple-action}, the logical connective negation is applied over the predicate \verb|(opened ?d)| to indicate that the expression \verb|(not (opened ?d))| is true when the door is closed. The predicates \verb|(door ?d)| and \verb|(not (opened ?d))| are linked using conjunction. The effects define how the state would change after the action application. The basic effect formula may include a predicate (\ie~an atom added to a state), a negation of predicate (\ie~an atom deleted from a state), or a conjunction of those. More complex effect formulas may include conditional effects and universally quantified effects. Note that predicates involved in actions are determined by the effects that actions have on the predicates, and the truthfulness of ground atoms included in the initial state of the problem description.

\subsection{Problem Description}
A problem description is specified with respect to a domain description. Each problem description begins with the declaration

\medskip
\verb|(define (problem <name>)(:domain <name>))|,

\medskip
\noindent which defines the name of the problem with respect to some application domain. For example, a problem description for a smart home can begin with the declaration 

\medskip
\verb|(define (problem instance1) (:domain smart-home))|.

\medskip
\noindent The simplest problem description specifies an initial state and a goal to be achieved. The description of the initial state is a list of all ground atoms that are true in the initial state. Recall that ground atoms are those predicates whose arguments are objects and constants, but not parameters. All other atoms are false due to the closed-world assumption. Fig.~\ref{fig:pddl-initial-state} shows a list of ground atoms describing an initial state and specified within the \verb|:init| field. It gives information about the layout of the home, available devices, their arrangement within the home, the statuses of devices, and other relations and properties. 

\begin{figure}[!ht]
\begin{tabular}{l}
\verb|(:init|\\
\hspace{0.5cm}\verb|(room livingRoom)|\\
\hspace{0.5cm}\verb|(room hallway)|\\
\hspace{0.5cm}\verb|(room bathroom)|\\
\hspace{0.5cm}\verb|(door d1)|\\
\hspace{0.5cm}\verb|(door d2)|\\
\hspace{0.5cm}\verb|(door d3)|\\
\hspace{0.5cm}\verb|(key k1)|\\
\hspace{0.5cm}\verb|(key k2)|\\
\hspace{0.5cm}\verb|(key k3)|\\
\hspace{0.5cm}\verb|(key k4)|\\
\hspace{0.5cm}\verb|(agent a1)|\\
\hspace{0.5cm}\verb|(adjacent livingRoom hallway)|\\
\hspace{0.5cm}\verb|(adjacent hallway bathroom)|\\
\hspace{0.5cm}\verb|(installed-in d1 hallway)|\\
\hspace{0.5cm}\verb|(installed-in d2 livingRoom)|\\
\hspace{0.5cm}\verb|(installed-in d3 bathroom)|\\
\hspace{0.5cm}\verb|(in a1 hallway)|\\
\hspace{0.5cm}\verb|(opened d2)|\\
\hspace{0.5cm}\verb|(locked d1)|\\
\hspace{0.5cm}\verb|(owns a1 k1)|\\
\hspace{0.5cm}\verb|(owns a1 k2)|\\
\hspace{0.5cm}\verb|(owns a1 k3)|\\
\verb|)|\\
\end{tabular}
\caption{PDDL ground atoms representing an initial state of the smart home in the key example from Figure~\ref{ch2aip:fig:ex1}.}\label{fig:pddl-initial-state}
\end{figure}

The goal description usually has the same form as the condition formula in action preconditions except that all predicates are ground. The goal formula is specified within a \verb|:goal| field.

\subsection{Types}
The domain description can be extended with other constructs, such as types, constants that have the same meaning for all planning problems in the given domain, functions to access and update numeric values (numeric fluents), and derived predicates. Types, for instance, are a syntactic sugar and can be useful to classify and represent various entities that share some characteristics into a more concise and human-readable form. Consider that the keys \verb|k1|, \verb|k2|, \verb|k3|, and \verb|k4| listed in Fig.~\ref{fig:pddl-initial-state} are of different types. For example, \verb|k1| is a digital key, \verb|k2| and \verb|k3| are pin tumbler lock keys (Yale keys), and \verb|k4| is car key. Since they are all keys, they can be represented in the domain description by using the \verb|key| type as shown in Fig.~\ref{fig:pddl-types}. All domains include \verb|object| and \verb|number| as built-in types, which usually reside at the top of the type classification. PDDL types are specified within the \verb|:types| field.

\begin{figure}[!ht]
\begin{tabular}{l}
\verb|(:types|\\
\hspace{0.5cm}\verb|key door room agent - object|\\
\hspace{0.5cm}\verb|digital-key yale-key car-key - key|\\
\verb|)|
\end{tabular}
\caption{PDDL types representing classes of smart home entities.}\label{fig:pddl-types}
\end{figure}

When types are provided in the domain description, the typing of parameters can be employed. This means that we can declare a type of argument of a predicate or a type of parameter of an action using the form 

\medskip
\verb|?x - TYPE-OF-X|. 

\medskip\noindent For example, \verb|(in ?a - agent ?r - room)|. Fig.~\ref{fig:pddl-typed-predicates} shows typed definitions of the predicates from Fig.~\ref{fig:pddl-simple-predicates}, and Fig.~\ref{fig:pddl-typed-action} illustrates a typed definition of the action schema from Fig.~\ref{fig:pddl-simple-action}.

\begin{figure}[!ht]
\begin{tabular}{l}
\verb|(:predicates|\\
\hspace{0.5cm}\verb|(adjacent ?r1 ?r2 - room)|\\
\hspace{0.5cm}\verb|(installed-in ?d - door ?r - room)|\\
\hspace{0.5cm}\verb|(in ?a - agent ?r - room)|\\
\hspace{0.5cm}\verb|(opened ?d - door)|\\
\hspace{0.5cm}\verb|(owns ?a - agent ?k - key)|\\
\verb|)|
\end{tabular}
\caption{PDDL typed predicates.}\label{fig:pddl-typed-predicates}
\end{figure}

\begin{figure}[!ht]
\begin{tabular}{l}
\verb|(:action open-door|\\
\hspace{0.5cm} \verb|:parameters (?d - door)|\\
\hspace{0.5cm} \verb|:precondition (and (not (opened ?d)))|\\
\hspace{0.5cm} \verb|:effect (opened ?d)|\\
\verb|)|\\
\end{tabular}
\caption{PDDL action schema with a typed parameter.}\label{fig:pddl-typed-action}
\end{figure}

\medskip\medskip\noindent
Consider now the problem description. In Fig.~\ref{fig:pddl-initial-state}, predicates are used to describe objects, for example, \verb|(room livingRoom)|. As a syntactic sugar, it is possible to describe such objects outside of the \verb|:init| field by using the \verb|:objects| field. In this case, the list of objects can be provided without or with types, making the problem description more concise (see Fig.~\ref{fig:pddl-initial-state-with-objects}).

\begin{figure}[!ht]
\begin{tabular}{l}
\verb|(:objects|\\
\hspace{0.5cm}\verb|livingRoom hallway bathroom - room|\\
\hspace{0.5cm}\verb|d1 d2 d3 - door|\\
\hspace{0.5cm}\verb|k1 - digital-key|\\
\hspace{0.5cm}\verb|k2 k3 - yale-key|\\
\hspace{0.5cm}\verb|k4 - car-key|\\
\hspace{0.5cm}\verb|a1 - agent|\\
\verb|)|\\
\verb|(:init|\\
\hspace{0.5cm}\verb|(adjacent livingRoom hallway)|\\
\hspace{0.5cm}\verb|(adjacent hallway bathroom)|\\
\hspace{0.5cm}\verb|(installed-in d1 hallway)|\\
\hspace{0.5cm}\verb|(installed-in d2 livingRoom)|\\
\hspace{0.5cm}\verb|...|\\
\verb|)|\\
\end{tabular}
\caption{More concise version of the problem description shown in Fig.~\ref{fig:pddl-initial-state} when objects are declared as an alternative to using predicates.}\label{fig:pddl-initial-state-with-objects}
\end{figure}

\subsection{STRIPS subset}
PDDL provides a STRIPS subset by which STRIPS planning problems can be represented in first-order logic with predicates, variables, and constants (for STRIPS planning problems, see Section~\ref{sec3}). Each action from a state model is represented using a STRIPS operator with preconditions and effects. Preconditions specify when an action is applicable and the effects describe what exactly is affected by the action. In the STRIPS representation, the effects are divided into an add list, which includes atoms to be made true in the successor state, and a delete list, which includes atoms to be made false. All atoms not specified in the representation of an operator remain unmodified. 

\newpage

\bibliographystyle{IEEEtran}
\bibliography{lit}

\end{document}